\documentclass[12pt,twoside,a4paper,reqno]{amsart}
\usepackage{bulletinUPB}
\usepackage{graphics}
\usepackage{amssymb}
\usepackage{eucal}
\usepackage{amsmath}
\usepackage{msc}
\usepackage{url}
\usepackage{booktabs}
\usepackage{multirow}
\usepackage{xcolor}

\usepackage[utf8]{inputenc}
\usepackage{epstopdf}
\usepackage{hyperref}

\title{Quantifying the Synthetic and Real Domain Gap in Aerial Scene Understanding}

\author{Alina MARCU$^1$}
\address{$^1$PhD Student, "Simion Stoilow" Institute of Mathematics of the Romanian Academy, e-mail: {\tt alina.marcu@upb.ro}}

\begin{document}

\pagestyle{headings}
\maketitle

\begin{abstract}
{\it Quantifying the gap between synthetic and real-world imagery is essential for improving both transformer-based models -- that rely on large volumes of data -- and datasets, especially in underexplored domains like aerial scene understanding where the potential impact is significant. This paper introduces a novel methodology for scene complexity assessment using Multi-Model Consensus Metric (MMCM) and depth-based structural metrics, enabling a robust evaluation of perceptual and structural disparities between domains. Our experimental analysis, utilizing real-world (Dronescapes) and synthetic (Skyscenes) datasets, demonstrates that real-world scenes generally exhibit higher consensus among state-of-the-art vision transformers, while synthetic scenes show greater variability and challenge model adaptability. The results underline the inherent complexities and domain gaps, emphasizing the need for enhanced simulation fidelity and model generalization. This work provides critical insights into the interplay between domain characteristics and model performance, offering a pathway for improved domain adaptation strategies in aerial scene understanding.}
\end{abstract}



\section{Introduction}
\label{sec:intro}
Robotics aims to develop physical agents capable of interacting with the real world, where vision plays a crucial role in perception and scene understanding. Advanced artificial intelligence (AI) systems depend on various learning paradigms. These include data-driven approaches (supervised or unsupervised) and experience-based methods (reinforcement learning). However, such systems encounter significant challenges in unstructured real-world environments. While the former relies on vast, high-quality datasets, the latter requires effective onboard computation, particularly for drones with strict size, weight, and power constraints.

Autonomous driving systems benefit from structured data collected by millions of vehicles, while aerial robots lack such infrastructure, making large-scale data collection more difficult. Although progress has been made in aerial scene understanding, the field is still far behind its ground-level counterpart. Autonomous vehicles have been extensively studied, resulting in well-defined benchmarks and methodologies~\cite{guo2021survey}, while aerial systems have seen comparatively fewer advancements~\cite{osco2021review}.

\begin{figure}[!ht]
\centering
\includegraphics[width=\linewidth,keepaspectratio]{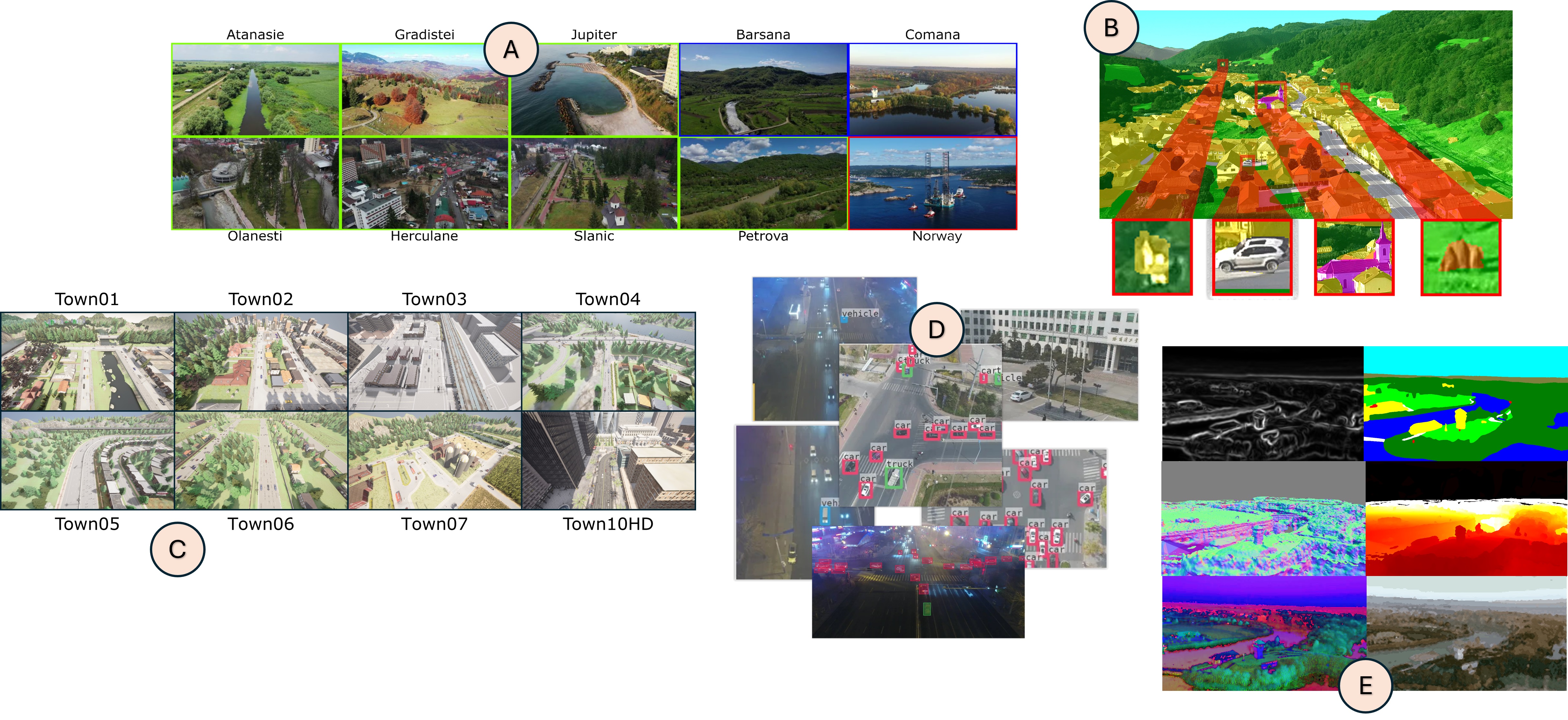}
\caption{Overview of the main challenges for aerial scene understanding: A) Environmental and altitude variations~\cite{marcu2023self}, B) Object scale variations within a frame~\cite{marcu2020semantics}, C) Discrepancies between unconstrained (real) and controlled (synthetic) environments~\cite{khose2023skyscenes}, D) Complex scene interpretation due to varying viewpoints, challenging lighting conditions, and occlusions in high-density areas~\cite{du2018unmanned}, E) Integration of multiple complementary scene representations for comprehensive understanding~\cite{marcu2023self}.}
\label{fig:0_main_challenges} 
\end{figure}

The unique perspectives and mobility of drones open opportunities for applications ranging from agriculture and infrastructure inspection to emergency response and urban planning~\cite{menouar2017uav}. However, these advantages also introduce distinct challenges in visual scene understanding (Figure~\ref{fig:0_main_challenges}), including viewpoint variability and the lack of large-scale annotated datasets. 

A major hurdle is the reliance on synthetic data generation to address the scarcity of annotated real-world aerial imagery. This raises a key question: How effectively do synthetic datasets represent the complexity and nuances of real-world scenes? Moreover, state-of-the-art vision transformer models~\cite{Dosovitskiy2020AnII}, typically trained on ground-level benchmarks~\cite{jain2023oneformer}, require adaptation to aerial domains, yet current evaluation practices often depend on time-consuming and error-prone manual annotations. To overcome these limitations, there is a need for unsupervised metrics that assess model adaptability before committing resources to data annotation. Such metrics can guide the effective use of synthetic and real datasets, ensuring meaningful advancements in model performance and dataset design.

In this paper, we address these gaps and contribute to aerial scene understanding and domain adaptation through the following:
\vspace{0.5em}
\begin{itemize}
    \item We propose Multi-Model Consensus Metric (MMCM), a novel, model-agnostic unsupervised metric for assessing perceptual complexity in aerial imagery. MMCM provides a quantitative evaluation of scene complexity based on the agreement of multiple state-of-the-art vision transformer models designed for semantic segmentation, using both agreement and confidence and eliminating the need for ground truth data.
    \item By integrating MMCM with structural metrics derived from depth estimation (e.g., depth entropy, discontinuity ratio), we offer a comprehensive approach for analyzing the perceptual and structural scene complexity. 
    \item We study the perceptual and structural gaps between synthetic (Skyscenes) and real-world (Dronescapes) datasets. We highlight the limitations of current synthetic datasets in representing real-world challenges, as well as their variability and potential for improvement.
    \item Our findings reveal how scene characteristics (e.g., depth discontinuities, perceptual complexity) affect the performance of state-of-the-art vision transformers, which highlights the need for increasing the robustness of such methods to address cross-domain variability.
\end{itemize}

\section{Aerial Scene Understanding}
\label{sec:aerial}
Scene understanding involves a broad range of tasks that can be categorized into five key domains: (1) object-centric tasks, focusing on the detection, classification, and tracking of individual entities within the scene; (2) semantic tasks, involving the interpretation of scene regions and their functional meanings; (3) geometric tasks, addressing the spatial layout and 3D structure reconstruction; (4) relational tasks, analyzing the interactions and relationships between scene elements; and (5) temporal tasks, dealing with the dynamic aspects and evolution of scenes over time. In drone-based visual scene understanding, two primary environments present unique challenges: indoor scenes, characterized by dense semantic content within confined spaces, and outdoor scenes, defined by vast scale variations and dynamic elements. This work focuses on outdoor environments, as they represent the primary operational domain for most UAV applications.

Much of the progress in scene understanding has been driven by advancements in autonomous driving, resulting in a terrestrial-centric approach. Consequently, the majority of datasets, methods, and technological innovations have been designed for ground-level perspectives~\cite{cordts2016cityscapes, geiger2013vision}, leaving aerial scene understanding comparatively underexplored. Recent trends in UAV-based datasets, as highlighted in Table~\ref{tab:table_1_general_comparison}, reveal the prevalence of synthetic datasets due to their ease of acquisition and the ability to generate noise-free representations that facilitate tackling challenging tasks. In contrast, real-world datasets focus on singular-tasks, such as semantic segmentation or simpler ones like object detection, reflecting the difficulties in acquiring high-quality annotations for multiple dense prediction tasks. In addition to semantic segmentation, depth estimation is another crucial task for scene understanding. This reliance on synthetic data, while pragmatic, underscores a critical gap: the scarcity of diverse and comprehensive real-world datasets limits UAV applications in complex, real-world scenarios. Promising efforts, such as the Dronescapes dataset~\cite{marcu2023self}, aim to address this by offering annotated real-world aerial imagery suitable for multiple advanced tasks.

A critical challenge in robotics is the \textbf{sim-to-real gap}, which refers to the performance disparity of models trained in simulated environments when deployed in real-world settings. This gap arises from the complexity of real-world scenarios, sensor limitations, computational constraints, and the dynamic nature of outdoor environments. While techniques such as domain randomization, transfer learning, and hybrid approaches have been proposed to bridge this gap, it remains a significant barrier to deploying robust systems.

In the aerial domain, this challenge is particularly pronounced due to the domain-specific characteristics of UAV operations. Analyzing the synthetic-to-real domain gap in aerial scene understanding is essential for identifying limitations in existing datasets, improving synthetic dataset design, and guiding model development to enhance real-world applicability. By addressing this gap, progress in aerial scene understanding can lead to UAVs that perform more consistently and reliably, particularly in real-world environments where such advancements are most critically needed.

\begingroup
    \setlength{\tabcolsep}{4pt}
    \begin{table}[!ht]
        \centering
        \caption{Overview of public UAV datasets throughout from 2023 and 2024, for both urban and rural scene understanding and of various types based on the environments of the data collection (real - R, synthetic - S, or both). Perspective denotes the UAV pitch angle ($\theta$) during image capture, such that F is forward view ($\theta = 0^\circ$), O is oblique view ($\theta \in (0^\circ, 90^\circ)$) and N is nadir view ($\theta = 90^\circ$). Magnitude denotes the scale of the dataset in the number of images / frames. "-" denotes missing information in the paper.}
        \label{tab:table_1_general_comparison}
        \resizebox{\textwidth}{!}{
            \begin{tabular}{lllcccccccc}
                \toprule
                 & \textbf{Year} & \textbf{Dataset} & \textbf{Type} & \textbf{Format} & \textbf{Task} & \textbf{Sensor} & \textbf{Altitude} & \textbf{Perspective} & \textbf{Resolution} & \textbf{Magnitude} \\
                 \cmidrule{2-11}
                \cmidrule{2-11}
                1 & 2023 & Dronescapes~\cite{marcu2023self} & R & Videos & \shortstack{Multi-task} & Multimodal & 50 - 70m & O & 3840 $\times$ 2160 & $\approx$90k \\
                2 &  & HIT-UAV~\cite{suo2023hit} & R & Images & \shortstack{Object \\ Detection} & Multimodal & 60 - 130m & O & 640 $\times$ 512 & $\approx$43k \\
                3 &  & Skyscenes~\cite{khose2023skyscenes} & S & Images & \shortstack{Multi-task} & Multimodal & 15, 35, 60m & (F, O, N) & 2160 $\times$ 1440 & 33.6k \\
                4 &  & SynDrone~\cite{rizzoli2023syndrone} & S & Images & \shortstack{Semantic \\ Segmentation} & Multimodal & 20, 50, 80m & (O, N) & 1920 $\times$ 1080 & 72k \\
                5 &  & UAVPal~\cite{maiti2023uavpal} & R & Images & \shortstack{Semantic \\ Segmentation} & RGB & 90m & N & 5472 $\times$ 3648 & $\approx$1.6k \\
                6 & & VDD~\cite{cai2023vdd} & R & Images & \shortstack{Semantic \\ Segmentation} & RGB & 50 - 120m & (O, N) & 4000 $\times$ 3000 & 400 \\
                \cmidrule{2-11}
                7 & 2024 & CART~\cite{lee2024cart} & R & Images & \shortstack{Semantic \\ Segmentation} & Multimodal & 40 - 120m & (O, N) & 960 $\times$ 600 & - \\
                8 &  & DDOS~\cite{Kolbeinsson_2024_CVPR} & S & Images & Multi-task & RGB & 1 - 25m & (F, O) & 1280 $\times$ 720 & 34k \\
                9 &  & WildUAV~\cite{blaga2024forest} & (R, S) & Images & Multi-task & Multimodal & 30, 50, 80m & (F, O, N) & - & 34k \\
                10 &  & UEMM-Air~\cite{liu2024uemm} & S & Images & \shortstack{Object \\ Detection} & Multimodal & 5 - 50m & (F, O, N) & 1920 $\times$ 1080 & $\approx$20k \\
                \bottomrule
                    
            \end{tabular}
        }
    \end{table}
\endgroup

\section{Multi-Model Consensus Metric (MMCM)}
\label{sec:method}
We propose a novel approach to measure the complexity of a scene, characterized by a set of frames/images, through the lens of pretrained vision transformers, without the need of labeled data. Our methodology leverages both perceptual complexity, measured via multi-model consensus, and structural complexity, captured through depth-based metrics. The key insight is that scene complexity manifests in both how consistently models interpret a scene and how intricate its physical structure is. \\

\vspace{0.5em}
\noindent\textbf{Perceptual Complexity via Model Consensus.} Given an image $I$ and an ensemble of $N$ semantic segmentation models $\mathcal{M} = {M_1, M_2, ..., M_N}$, we obtain for each model $M_i$ a segmentation map $S_i(x,y) \in {1,...,K}$ where $K$ is the number of classes, and a confidence map $C_i(x,y) \in [0,1]$ derived from softmax probabilities. For any two models $M_i$ and $M_j$, we define their weighted agreement score $A_{i,j}$ as:
\begin{equation}
A_{i,j} = \frac{1}{|I|} \sum_{x,y} \delta(S_i(x,y), S_j(x,y)) \sqrt{C_i(x,y) C_j(x,y)}
\end{equation}
where $|I|$ is the number of pixels and $\delta(a,b)$ is the Kronecker delta function defined as:
\[ \delta(a,b) = \begin{cases} 1 & \text{if } a = b \\ 0 & \text{otherwise} \end{cases} \]
This function acts as a binary indicator of semantic agreement - it equals 1 when both models assign the same semantic class to a pixel, and 0 when they disagree. The mean agreement score $\bar{A}$ across all model pairs is:
\begin{equation}
\bar{A} = \frac{2}{N(N-1)} \sum_{i=1}^{N-1} \sum_{j=i+1}^N A_{i,j}
\end{equation}

and the mean confidence score $\bar{C}$ across all models is:
\begin{equation}
\bar{C} = \frac{1}{N} \sum_{i=1}^N \left(\frac{1}{|I|} \sum_{x,y} C_i(x,y)\right)
\end{equation}

The agreement is weighted by the geometric mean of the models' confidence at each pixel. This ensures that high agreement scores require two elements: consistent predictions and high confidence from both models. The consensus score $MMCM$ becomes the combination between the mean agreement across all model pairs with their average confidence:
\begin{equation}
MMCM(I) = \bar{A} \sqrt{\bar{C}}
\end{equation}

The consensus score exhibits several important mathematical properties. First, $MMCM(I) \in [0,1]$ for all images $I$, providing a normalized measure of complexity. A score of $MMCM(I) = 1$ indicates perfect perceptual alignment - complete agreement between all models with maximum confidence. Conversely, $MMCM(I) = 0$ represents maximum perceptual complexity, arising from either complete model disagreement or zero confidence in predictions. Therefore lower consensus translates to higher scene complexity due to its characteristics. 

The semantic consensus mechanism represents a significant advancement over traditional ensemble methods. Rather than simple majority voting or averaging, our approach implements a weighted consensus system that considers both inter-model agreement and confidence scores. This is achieved through a novel scoring function that combines geometric mean of confidence values with a spatial agreement coefficient, providing a more nuanced understanding of prediction reliability.

\vspace{0.5em}
\noindent\textbf{Cross-Domain Gap Analysis.} For dataset-level analysis, given a dataset $\mathcal{D}$ containing $M$ images, we compute the mean consensus score $\mu_{MMCM}$:
\begin{equation}
\mu_{MMCM} = \frac{1}{M} \sum_{I \in \mathcal{D}} MMCM(I)
\end{equation}

When comparing any two domains (either synthetic-synthetic, real-real or synthetic-real) or datasets $\mathcal{D}_1$ and $\mathcal{D}_2$, we define their relative perceptual gap $\rho_{PG}$ as:
\begin{equation}
\rho_{PG} = \frac{|\mu_{MMCM}(\mathcal{D}_1) - \mu_{MMCM}(\mathcal{D}_2)|}{\max(\mu_{MMCM}(\mathcal{D}_1), \mu_{MMCM}(\mathcal{D}_2))}
\label{eq:perceptual_gap}
\end{equation} \\
This formulation provides a symmetric, normalized measure of discrepancy between domains, where values closer to 0 indicate stronger alignment in terms of perceptual complexity or similarity. 

It is worth mentioning that MMCM leverages vision transformers that were pretrained on datasets predominantly consisting of ground-level natural images (as later discussed in Section~\ref{sec:experiments}). This might introduce biases when applied to aerial imagery. However, MMCM, as a model-agnostic metric, focuses on the agreement and confidence of predictions across multiple models, thus reducing reliance on the specific pretraining dataset. In this way we are able to distinguish between cases of shared pretraining biases versus genuine scene understanding. Additionally, by incorporating diverse models (Mask2Former~\cite{cheng2021mask2former}, OneFormer~\cite{jain2023oneformer}, SegFormer~\cite{xie2021segformer}), we reduce the impact of architecture-specific biases that could arise from similar pretraining strategies.

\vspace{0.5em}
\noindent\textbf{Structural Complexity via Depth Analysis.} We complement the perceptual analysis with metrics derived from monocular depth estimation. Given a depth map $D(x,y)$, we compute the depth entropy $H_D$ to capture the diversity of depth values:
\begin{equation}
H_D = -\sum_{i=1}^{B} p_i \log p_i
\end{equation}
where $p_i$ represents the probability of depth values in bin $i$ of $B$ total bins. Higher entropy indicates more varied depth distributions, typical of scenes with complex structural arrangements. 

Additionally, to quantify structural transitions in the depth map, we compute the depth gradients using the Sobel operators \(S_x\) and \(S_y\)~\cite{kanopoulos1988design}, which are convolution kernels designed to approximate the image gradient. The Sobel operator emphasizes changes in intensity (or depth values in our case) along the horizontal and vertical directions. The Sobel operator in the horizontal direction is defined as:
\[
S_x = \begin{bmatrix}
-1 & 0 & +1 \\
-2 & 0 & +2 \\
-1 & 0 & +1 \\
\end{bmatrix},
\]
and in the vertical direction as:
\[
S_y = \begin{bmatrix}
-1 & -2 & -1 \\
0 & 0 & 0 \\
+1 & +2 & +1 \\
\end{bmatrix}.
\]
These operators are convolved with the depth map \(D\) to compute the gradients in the x and y directions:
\[
G_x(x, y) = S_x * D, \quad G_y(x, y) = S_y * D,
\]
where '*' denotes the convolution operation. The magnitude of the depth gradient at each pixel is then calculated using:
\begin{equation}
G_D(x, y) = \sqrt{G_x(x, y)^2 + G_y(x, y)^2}.
\end{equation}
This gradient magnitude \(G_D(x, y)\) represents the rate of change in depth at each pixel, highlighting areas with significant structural transitions in the scene or regions with high gradient magnitudes correspond to depth discontinuities or edges, which implies greater scene complexity.

The discontinuity ratio $R_d$ measures the proportion of significant depth transitions relative to the total number of pixels:
\begin{equation}
R_d = \frac{1}{|I|} \sum_{x,y} \chi_{[G_D(x,y) > \tau \cdot (D_{max} - D_{min})]}
\end{equation}
where $\chi_{[condition]}$ is the characteristic function that equals 1 when the condition is true and 0 otherwise, $\tau$ is a relative threshold (typically set to 0.1), and $(D_{max} - D_{min})$ is the depth range in the scene. This adaptive thresholding ensures that discontinuities are identified relative to the scene's depth scale.


\section{Experimental Analysis}
\label{sec:experiments}
For our experimental analysis, we used two of the most recent and representative datasets that provide complementary perspectives for aerial scene understanding (synthetic~\cite{khose2023skyscenes} and real~\cite{marcu2023self} imagery).  

\vspace{0.5em}
\noindent\textbf{Datasets details.} The first dataset, Dronescapes~\cite{marcu2023self} consists of videos collected from real-world drone flights with a large variation in spatial distributions of classes among 10 different scenes, which range from rural (Atanasie, Gradistei, Petrova, Barsana, Comana), to urban (Olanesti, Herculane, Slanic) and seaside (Jupiter, Norway), while
also being geographically far apart. The videos were collected at altitudes between 40-70 meters with oblique views (pitch of $\approx$ 45 degrees). In our experiments we used the same sampling procedure proposed by the authors (when they manually annotated the frames for the semantic segmentation task). Frames are sampled at regular intervals, once every 2 seconds, which yields between 20-50 frames per scene, with a total of 299 frames used in our analysis for this dataset. 

The second dataset, Skyscenes~\cite{khose2023skyscenes} consists of synthetic aerial imagery generated to simulate drone perspectives across diverse environments. The dataset encompasses both urban and rural scenes (a total of 8 scenes with 70 frames for each), captured under varying conditions including different weather patterns, times of day, and viewing angles. To ensure a fair comparison, we opted to use the scenes captured under clear noon lighting conditions under similar capturing conditions to the real dataset (altitude of 60 meters and a pitch of 45 degrees). To create a dataset similar in scale to the real one, we uniformly sampled 33 frames from each of the 8 scenes, resulting in a total of 264 images used in our experiments.

A closer examination reveals significant structural and semantic differences between the two datasets. First, Dronescapes offers a more diverse distribution of object scales, ranging from small objects like cars and people in urban settings to large vegetation areas in rural scenes, influenced by variations in flying altitudes (40–70m). In contrast, Skyscenes features more uniform object scales due to the procedural generation of synthetic environments and a fixed altitude of 60m. Second, lighting variations are more prominent in Dronescapes, as videos were captured at different times of day, introducing natural elements such as shadows, reflections, and diffused lighting. Skyscenes, however, simulates controlled conditions with uniform lighting under clear skies. Perhaps most notably, the datasets show distinct class distribution patterns. Dronescapes reflects real-world imbalances, with certain classes (vegetation, buildings) dominating specific scenes, while privacy regulations and data collection constraints result in underrepresenting others (such as cars, persons). Skyscenes, however, maintains more balanced class distributions through its controlled generation process, achieving better representation even for typically underrepresented classes, since data privacy constraints do not apply.
These disparities highlight critical areas for improving synthetic dataset generation, specifically the need to: (1) incorporate more diverse object scales and perspectives and varying altitudes, (2) simulate natural lighting conditions (with shadows, reflections and different weather conditions), and (3) reproduce realistic class distribution patterns that better reflect real-world scenarios while maintaining representation of minority classes.


\vspace{0.5em}
\noindent\textbf{Experimental setup.} MMCM is model-agnostic and can be computed for any number of models. To establish the scene-wise perceptual complexity, we used three state-of-the-art vision transformers models for semantic segmentation (Mask2Former~\cite{cheng2021mask2former}, OneFormer~\cite{jain2023oneformer}, SegFormer~\cite{xie2021segformer}). These models follow a similar training procedure -- pretrained on ImageNet~\cite{deng2009imagenet} and then finetuned on Cityscapes~\cite{cordts2016cityscapes} for the task of semantic segmentation with a fixed set of 19 classes that are mostly encountered in urban scenes (road, sidewalk, building, wall, fence, pole, traffic light, traffic sign, vegetation, terrain, sky, person, rider, car, truck, bus, train, motorcycle and bicycle), since Cityscapes is the most well-known benchmark for self-driving cars. By maintaining consistency in choosing the same dataset the models were finetuned on, we minimize the impact dataset biases might have, allowing us to attribute the differences in the perceptual complexity metric directly to the architectural variation of the models of choice. For assessing structural complexity, we utilized the most recent foundation model for depth estimation, DepthAnythingV2~\cite{yang2024depth}, known for its high accuracy and generalization capabilities across diverse scenes. For a fair comparison, we applied the same pre-processing pipeline to the images from both datasets and rescaled them to 960 $\times$ 540 before model predictions. 

\vspace{0.5em}
\noindent\textbf{Per-scene MMCM evaluation.} We report the mean MMCM per scene for both real and synthetic datasets in Table~\ref{tab:table_2_consensus_per_scene}. These results reveal higher mean values across scenes for data collected from real-world environments (0.6926) compared to synthetic ones (0.5693), which indicates stronger model consensus in real domains. These differences should be interpreted with consideration of the models' pretraining bias, since they were trained primarily using natural images (from real-world environments) although from a different perspective (ground-view vs. aerial-view). To minimize the impact of datasets biases in our experimental analysis, we ensured consistent pretraining and fine-tuning datasets (e.g., Cityscapes for semantic segmentation) across all models, allowing a fair comparison of their agreement rather than absolute performance. The comparative analysis between real-world and synthetic datasets highlights scenarios where pretrained biases may manifest (e.g., lower MMCM scores for synthetic scenes). However, the varying MMCM scores across different synthetic scenes (from 0.4852 to 0.7274) indicates that our metric is also sensitive to scene-specific characteristics beyond synthetic and real domains differences. These numbers not only reflect notable inter-domain gap, but also scene differences within a specific dataset, which we will also show later on in our experiments.

We argue that in scenes with high MMCM scores, models show consistent predictions suggesting that these scenes are simpler, more homogeneous, or well-represented in the training data (e.g., static objects like roads, buildings, land or vegetation depending on the scene's characteristics - urban or rural). Conversely, scenes with low MMCM scores reveal substantial variability in predictions, likely driven by challenging environmental factors such as diverse object scales, intricate textures, or lighting variations, which strain the models' ability to achieve consensus, through consistent predictions.

\begingroup
    \setlength{\tabcolsep}{4pt}
    \begin{table*}
        \centering
        \caption{MMCM comparison for both real and synthetic datasets. The numbers reported are averaged per scene. We can see that metric is higher on the dataset collected from a real-world environment meaning that the models have a higher agreement and confidence, making these scenes easier to segment mainly due to the fact that the models were also trained using natural images.}
        \label{tab:table_2_consensus_per_scene}
        \resizebox{\textwidth}{!}{
            \begin{tabular}{ccccccccccc}
                \toprule
                \multicolumn{11}{c}{\textbf{Dronescapes (real-world environment)}} \\
                \midrule
                Atanasie & Barsana & Comana & Gradistei & Herculane & Jupiter & Norce/Norway & Olanesti & Petrova & Slanic & Mean \\
                0.6953 & 0.7201 & 0.6879 & 0.7761 & 0.6765 & 0.6229 & 0.7516 & 0.5869 & 0.7256 & 0.6827 & 0.6926 \\
                \midrule
                \multicolumn{11}{c}{\textbf{Skyscenes (synthetic, rendered environment)}} \\
                \midrule
                \phantom{text} & Town01 & Town02 & Town03 & Town04 & Town05 & Town06 & Town07 & Town10HD & \phantom{text} & Mean \\
                \phantom{text} & 0.5504 & 0.7274 & 0.6108 & 0.4857 & 0.4852 & 0.4877 & 0.5446 & 0.6624 & \phantom{text} & 0.5693 \\
                \bottomrule
            \end{tabular}
        }
    \end{table*}
\endgroup

\begin{figure*}[!ht]
\centering
\includegraphics[width=\linewidth,keepaspectratio]{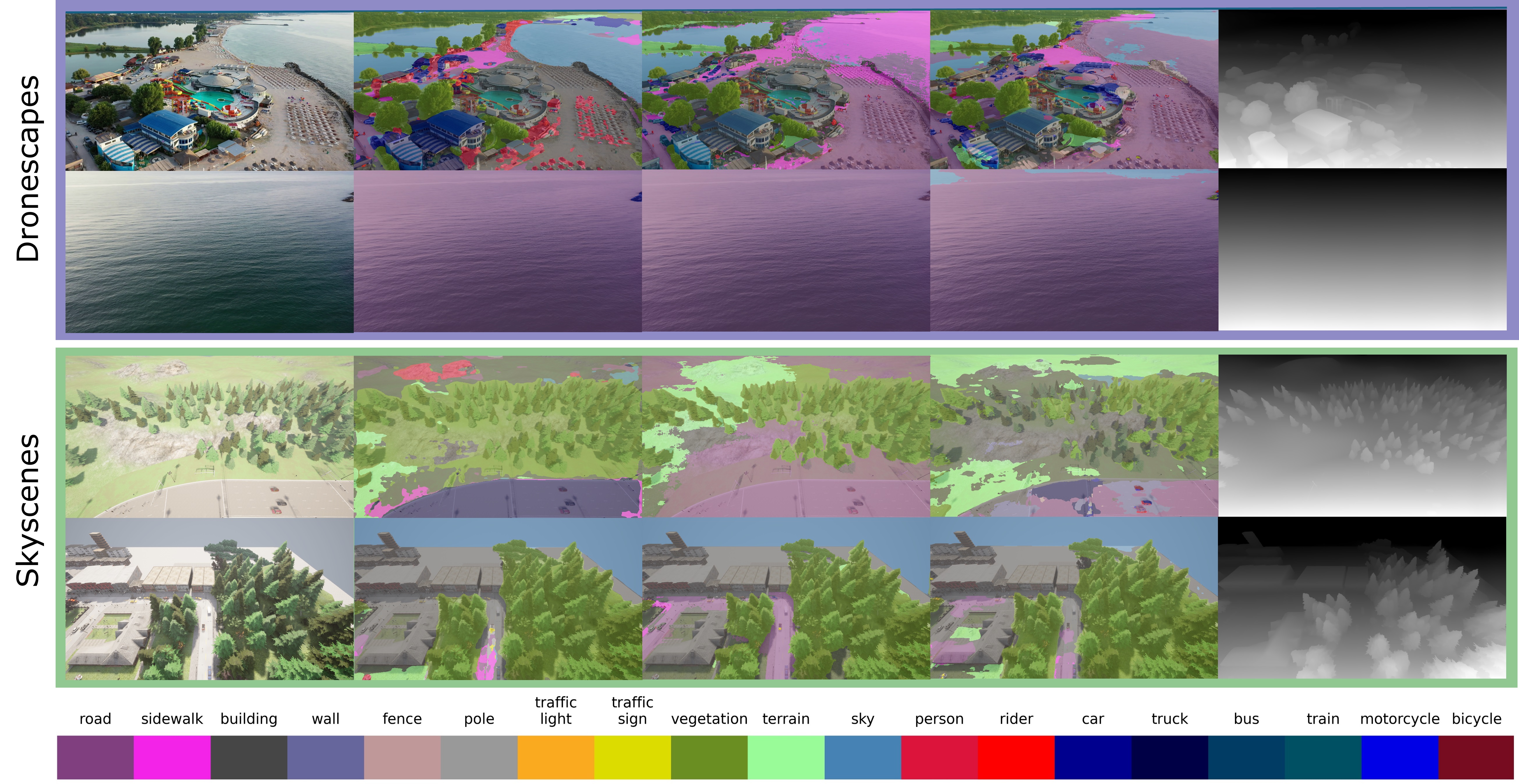}
\caption{Qualitative evaluation of the images with the lowest and highest MMCM score for each dataset (Dronescapes - real and Skyscenes - synthetic). From left to right in order we present the RGB image, semantic segmentation predictions from Mask2Former, OneFormer, SegFormer and depth prediction normalized from DepthAnythingV2. On the bottom of the figure the legend for the Cityscapes classes.}
\label{fig:3_qualitative_results}
\end{figure*}

\vspace{0.5em}
\noindent\textbf{Qualitative evaluation.} Our quantitative evaluation is also backed up by qualitative results in Figure~\ref{fig:3_qualitative_results} which highlights the frames with the lowest and highest MMCM scores from each dataset. The first row illustrates a complex coastal scene from Dronescapes where the three models (Mask2Former, OneFormer, and SegFormer) demonstrate notable disagreement in their predictions, particularly in challenging areas such as the beach-water boundary and fine architectural details. In contrast, the second row from Dronescapes shows a water surface scene, where all models achieve high agreement in their predictions, consistently classifying the uniform water surface. This aligns with our earlier observation that simpler scenes tend to yield higher model consensus. For Skyscenes, both scenes contain similar elements (vegetation, buildings, roads). However, we observe varying levels of model agreement. The more complex scene (top) shows disagreement in predictions, particularly at object boundaries and in areas with multiple overlapping elements. In contrast, the simpler synthetic scene (bottom) demonstrates more consistent predictions. This consistency is achieved because the objects are well-defined by texture, especially buildings and vegetation.


A key advantage of MMCM is that it operates independently of ground truth data. It serves as a measure of model reliability and consistency rather than performance. However, the first qualitative example on Dronescapes highlights a critical limitation. MMCM fails to account for systematic errors rooted in dataset bias. The models consistently misclassify the water surface as a road. This stems from multiple factors: (1) the models' closed-set training on Cityscapes, where water surfaces are not represented as a distinct class, (2) the visual similarity between reflective water surfaces and asphalt roads when viewed from above, particularly in terms of texture and color patterns, and (3) the context-based biases in the models, where large flat surfaces in urban environments are predominantly roads. General approaches to address these limitations include: (1) fine-tuning the models on aerial datasets that include water bodies as a distinct class would help establish proper class boundaries, (2) and also incorporating multi-modal data such as near-infrared bands or depth information could provide additional discriminative features, as water bodies have distinctive spectral and depth signatures compared to roads or (3) addressing the problem of domain adaptation from an aerial-specific viewpoint using recent and advanced open-set methods for semantic segmentation~\cite{choe2024open}.

\begin{figure*}[!ht]
\centering
\includegraphics[width=\linewidth,keepaspectratio]{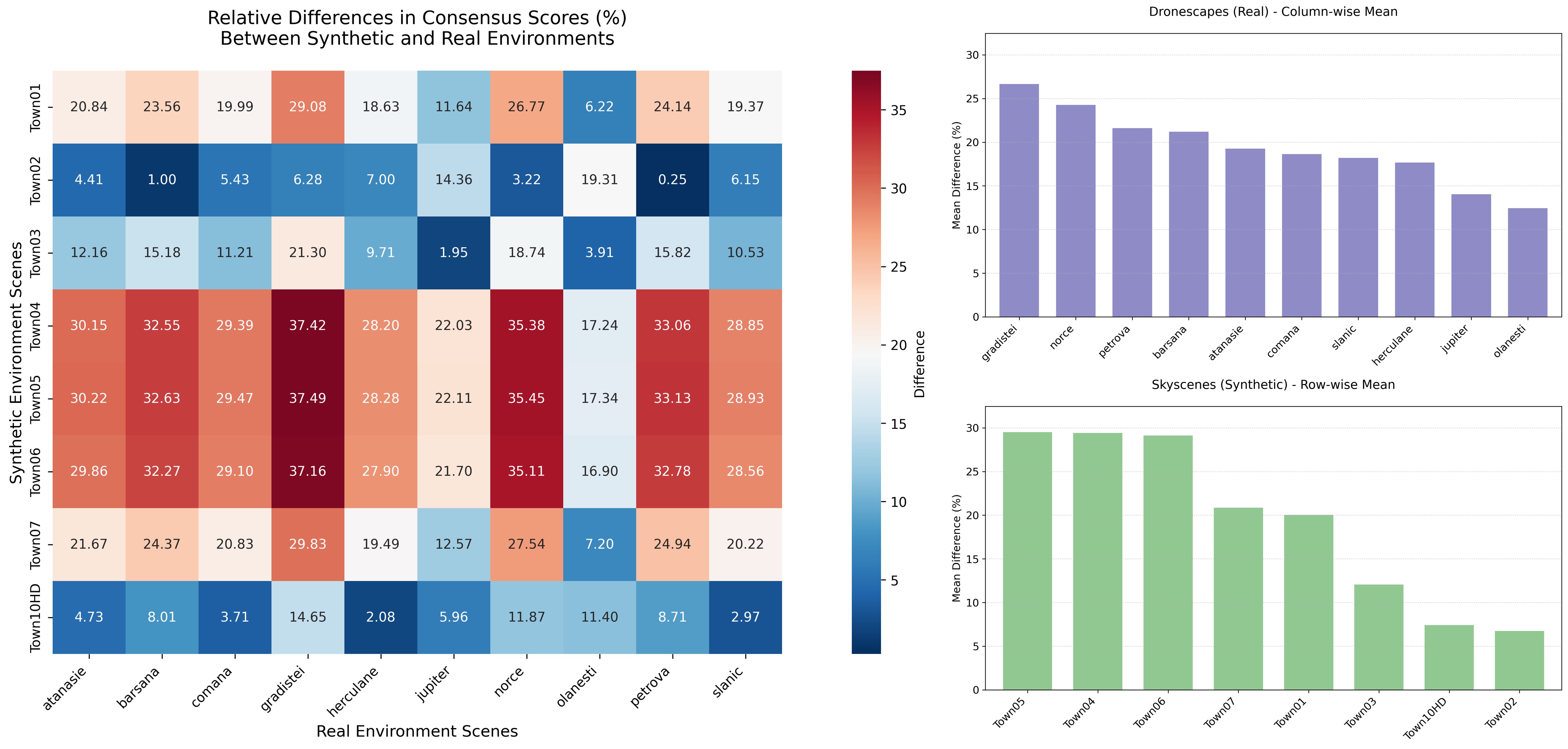}
\caption{(Left) Cross-domain relative perceptual gap ($\rho_{PG}$) between real and synthetic environments. (Right) Mean aggregated relative differences (derived from the heatmap on the left-side of the figure).}
\label{fig:1_inter_dataset_analysis}
\end{figure*}

\vspace{0.5em}
\noindent\textbf{Inter-domain MMCM analysis.} We measure the relative perceptual gap (Equation~\ref{eq:perceptual_gap}) between each real and synthetic scene from the datasets. On the left-side of Figure~\ref{fig:1_inter_dataset_analysis}, the heatmap reveals distinct patterns in the relative differences of the consensus scores. Most notably, synthetic environments Town04, Town05, and Town06 consistently show the highest relative differences (30-37\%) across all real environment scenes, reaching the maximum values when each of them are compared to the Gradistei scene. We can infer that these scenes present significantly more challenging scenarios for model consensus compared to real scenes. On the other hand, Town02 and Town10HD exhibit markedly lower relative differences across most real scenes, which indicates that their characteristics more closely align with real-world environments in terms of model consensus. This alignment is particularly evident in comparisons with Olanesti (6.22\% for Town01) and Herculane (2.08\% for Town10HD), suggesting these synthetic-real pairs share similar complexity patterns, but the best alignment is between Town02 with Petrova of (0.25\% difference). A notable pattern across real scenes is that Gradistei, Norce and Petrova consistently show the highest relative differences across most synthetic environments, but in particular with Town04, Town05 and Town06. This suggests that these real scenes may possess inherent characteristics (unique elements within the scene) that systematically influence model consensus, regardless of the synthetic environment used for comparison. These patterns are easily visible in the aggregated relative differences on the right-side of Figure~\ref{fig:1_inter_dataset_analysis}, computed separately for each dataset and arranged in descending order. Gradistei exhibits the highest mean difference ($\approx$27\%) across all synthetic comparisons, while Olanesti shows the lowest ($\approx$12\%). Also from the synthetic environment, scenes Town04, Town05, and Town06 exhibit significantly high mean differences ($\approx$29\%). This clustering of high differences among similar town environments suggests that these synthetic scenes may share common characteristics that consistently challenge model consensus. Conversely, Town02 and Town10HD show substantially lower mean differences ($\approx$7\%), indicating their visual characteristics may better approximate real-world conditions. The high range of differences (from $\approx$29\% to $\approx$7\%) demonstrate varying degrees of domain gap between synthetic and real environments, with some synthetic scenes more successfully approximating the complexity characteristics of real-world scenarios than others. 

\begin{figure*}[!ht]
\centering
\includegraphics[width=\linewidth,keepaspectratio]{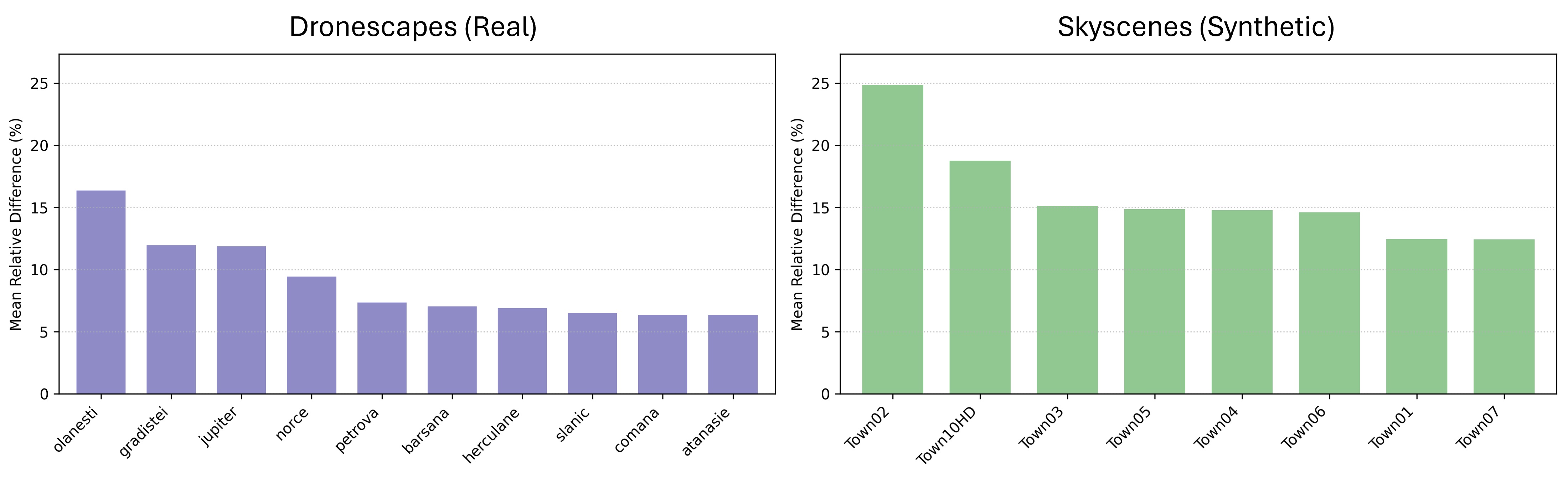}
\caption{Evaluating the relative perceptual gap within each dataset (mean aggregated for each scene in the dataset) for the real environment - Dronescapes (left) and synthetic one - Skyscenes (right).}
\label{fig:2_intra_dataset_analysis}
\end{figure*}

\vspace{0.5em}
\noindent\textbf{Intra-domain MMCM analysis.} An interesting aspect of measuring the relative perceptual gap is that it can be computed between any two scenes, therefore we can derive some interesting insights by exploring the intra-dataset differences and determining the most similar or dissimilar scenes in terms of model consensus. We present such an analysis, showing the variance within both real and synthetic environments in Figure~\ref{fig:2_intra_dataset_analysis}. For Dronescapes, we observe that scenes such as Olanesti and Gradistei exhibit higher relative differences in consensus metrics, indicating significant variability in model agreement, likely due to complex features, such as challenging lighting conditions, dramatic object scale differences or diverse scene compositions, which strain model consistency. Also, based on the information provided by the authors~\cite{marcu2023self}, Olanesti represents an urban layout, while Gradistei portrays a rural landscape. On the other hand, scenes like Atanasie and Comana demonstrate minimal variation, pointing to more uniform or predictable model behavior. This is likely due to simpler scene layout or more consistent scene characteristics. Atanasie's lower-altitude captures predominantly focus on vegetation and land features, while Comana's high-altitude perspective diminishes fine-grained details, resulting in scenes dominated by well-represented static object classes that models handle more consistently.




For Skyscenes, the highest relative differences are observed in Town02, followed by Town10HD, reflecting greater variability in model agreement. This is attributed to the complexity or richness of these synthetic scenes, since they reflect mostly industrial or high urbanistic views with tall buildings and very sparse vegetation compared to the other scenes. However, the differences between them are significant, indicating distinct types of dissimilarities: Town10HD primarily features clear views of man-made structures characteristic of metropolitan areas, such as buildings, while Town02 presents cluttered scenes with a wide variety of elements—a mix of small man-made constructions and natural elements. Scenes like Town07 and Town01 exhibit lower relative differences, indicating a more stable and uniform model response, since they have very few fine-grained objects, resembling a rural environment - sparse man-made constructions and mostly well-known vegetation, land and a few trees, which denotes simplicity. The most tight cluster is formed by Town04, Town05, and Town06 suggesting strong similarities between these scenes (or high redundancy due to sharing overlapping views).

The results also indicate that the Skyscenes dataset exhibits greater variability compared to Dronescapes, as reflected by the consistently larger relative differences in the synthetic scenes. This highlights a potential gap between synthetic and real-world data, which can be interpreted in two ways. First, it underscores the need for enhanced simulation fidelity to better align synthetic data with real-world conditions. Alternatively, the higher variability in Skyscenes could be seen as a strength, offering a broader range of scenery. This diversity in Skyscenes might also reveal Dronescapes' biases towards a specific type of regions.

\begin{figure*}[!ht]
\centering
\includegraphics[width=\linewidth,keepaspectratio]{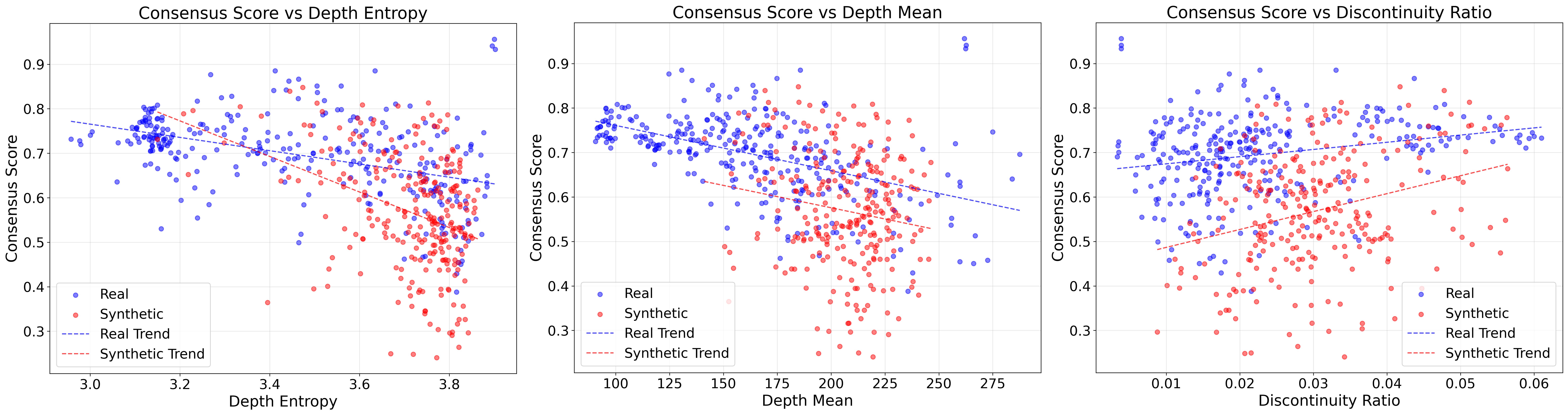}
\caption{Analysis of the perceptual and structural scene complexity comparison between samples from real and synthetic datasets. (Left) MMCM vs. Depth Entropy. (Middle) MMCM vs. Depth Mean. (Right) MMCM vs. Discontinuity Ratio. The dotted lines represent the linear trend (linear regression fit) between the metrics on the x-axis and the consensus scores on the y-axis for both real and synthetic data to help visualize {\color{blue} the overall positive or negative correlation between the scores.}}
\label{fig:4_depth_complexity_analysis}
\end{figure*}

\vspace{0.5em}
\noindent\textbf{Perceptual and structural scene complexity comparison.} Our analysis of depth entropy, mean depth, and discontinuity ratio compared to MMCM shown in Figure~\ref{fig:4_depth_complexity_analysis} reveals distinct patterns in how real and synthetic aerial imagery affect segmentation model agreement. We also show the performance discrepancy of the best pretrained vision transformer models for semantic segmentation and depth estimation when tackling novel aerial domains. Depth entropy shows clear distributional disparities between real and synthetic data. Real-world images exhibit stable consensus scores (0.6-0.8) while clustering in lower entropy ranges (3.0-3.4). In contrast, synthetic images exhibit higher entropy ranges (3.4-3.8) with a pronounced degradation in consensus as entropy increases, highly due to faulty depth estimation reflecting higher scene complexity or low adaptability of the depth model to the current testing scenario. Analyzing the mean depth (image-wise mean) provides a complementary perspective. Real images maintain relatively consistent consensus scores across depth means (values between 100-200, not meters), with a small negative correlation. Synthetic images, however, show a steeper decline in consensus with increasing depth means (175-250) and significantly higher variance in performance. Again this comparison indicates that the synthetic scenes challenge current segmentation models more severely at greater depths. The discontinuity ratio offers insights into how abrupt depth transitions affect segmentation reliability. Both datasets show positive correlations, but with notably different characteristics. Real images maintain more stable consensus scores across discontinuity ratios, while synthetic images exhibit much higher variance, particularly in mid-range values (0.02-0.04). This suggests that while both real and synthetic scenes benefit from clear depth boundaries, the discontinuity patterns from the synthetic scenes do not align with real-world scenarios, due to poor adaptability of the depth model to this scenario. 

\section{Conclusion}
\label{sec:conclusion}
This study highlights the importance of understanding the complexity of a scene in bridging the sim-to-real gap for comprehensive aerial scene understanding. By introducing MMCM, a model-agnostic metric for evaluating perceptual complexity, a multi-model consistency metric, and complementing it with depth-based structural metrics, we provide a comprehensive framework for analyzing intra and inter-domain disparities. Our results demonstrate that real-world datasets generally exhibit higher model consensus and structural stability, while synthetic datasets showcase greater variability and a pronounced need for alignment with real-world conditions. 

Our insights have direct implications for real-world UAV applications. The perceptual complexity metric (MMCM) combined with the structural complexity one can serve as a valuable tool for guiding UAV navigation algorithms to adaptively adjust their behavior in real-time based on the perceived complexity of the environment. For instance, areas identified as high complexity (or low MMCM scores) could trigger cautious navigation strategies, slower flight speeds, or increased reliance on complementary sensors for accurate decision-making. Regions showing consistent model predictions across multiple frames, coupled with favorable depth characteristics (low entropy and few discontinuities), could be automatically flagged as potential safe landing zones. Another example, would be obstacle avoidance using the depth confidence estimates - critical in scenarios where reliable distance estimation is essential for safe operation.

These contributions establish a foundation for advancing domain adaptation and generalization in aerial and other specialized domains. Future work will be focused on improving the complexity metrics using temporal information and exploring their integration into adaptive learning pipelines to further reduce domain disparities and improve the robustness of learned vision models in aerial scene understanding.


\bigskip


\bibliographystyle{abbrv}
\bibliography{paper}

%
%
%
%
%
%

\end{document}